\documentclass[letterpaper, 10 pt, conference]{ieeeconf}  

\IEEEoverridecommandlockouts                              

\usepackage{graphics} 
\usepackage{amsmath} 
\usepackage{amssymb}  
\usepackage{graphicx}
\usepackage{multirow}
\usepackage{color}

\definecolor{revised}{rgb}{1,0,0}

\title{\LARGE \bf
Robust LiDAR-Camera Calibration with 2D Gaussian Splatting
}

\author{Shuyi Zhou$^{1}$, Shuxiang Xie$^{1}$, Ryoichi Ishikawa$^{1}$, Takeshi Oishi$^{1}$
\thanks{This work was supported by JSPS KAKENHI Grant Numbers JP24K21173 and JP24H00351.}
\thanks{$^{1}$The authors are with The Institute of Industrial Science, The University of Tokyo, Japan. Emails:
        {\tt\small \{zhoushuyi495, shxxie, ishikawa,  oishi\}@cvl.iis.u-tokyo.ac.jp}}%
}

\DeclareMathOperator*{\argmin}{arg\,min}
\begin{document}

\maketitle

\begin{abstract}

LiDAR-camera systems have become increasingly popular in robotics recently. 
A critical and initial step in integrating the LiDAR and camera data is the calibration of the LiDAR-camera system. 
Most existing calibration methods rely on auxiliary target objects, which often involve complex manual operations, whereas targetless methods have yet to achieve practical effectiveness. 
Recognizing that 2D Gaussian Splatting (2DGS) can reconstruct geometric information from camera image sequences, we propose a calibration method that estimates LiDAR-camera extrinsic parameters using geometric constraints. 
The proposed method begins by reconstructing colorless 2DGS using LiDAR point clouds. 
Subsequently, we update the colors of the Gaussian splats by minimizing the photometric loss. 
The extrinsic parameters are optimized during this process. 
Additionally, we address the limitations of the photometric loss by incorporating the reprojection and triangulation losses, thereby enhancing the calibration robustness and accuracy.
 
\end{abstract}



\section{INTRODUCTION}
LiDAR-camera fusion plays a critical role in autonomous driving and robotics. 
By integrating accurate depth measurements from LiDAR with dense optical scans provided by cameras, we can develop robust solutions for various tasks, including object detection~\cite{yoo20203d}, simultaneous localization and mapping (SLAM)~\cite{zhong2021survey}, and 3D reconstruction~\cite{carlson2023cloner}. 
However, for effective sensor fusion, it is crucial to represent data from different sensors in a unified coordinate system.

Sensor calibration is an important preliminary step in integrating measurements from multiple sensors. 
The extrinsic calibration process attempts to determine the relative pose, encompassing translation and rotation between the sensors.
Calibration among different sensors is typically challenging caused by variations in the density, sensor modalities, field of view, and resolution. 
Therefore, most traditional LiDAR-camera calibration methods use auxiliary calibration target objects, such as textured plane objects~\cite{verma2019automatic, zhao2020extrinsic, sim2016indirect, yang2012simple}. 
However, these methods can be complex to set up and require laborious manual operations.

In the past decade, targetless calibration methods have emerged, beginning with techniques that use appearance correspondences, such as edges~\cite{levinson2013automatic, yuan2021pixel} and intensities~\cite{pandey2012automatic, koide2023general} between LiDAR and camera frames. 
These methods have been followed by approaches that leverage deep learning techniques 
\cite{iyer2018calibnet, lv2021lccnet}.
Geometric constraints have also proven effective~\cite{ishikawa2018lidar, ishikawa2024lidar}, particularly with 
dense 3D reconstruction from images~\cite{schoenberger2016mvs}. 
Recent advances in 3D reconstruction methods, such as neural radiance fields (NeRF)~\cite{mildenhall2020nerf} and Gaussian splattings (GS)~\cite{kerbl20233d}, have further refined this approach~\cite{zhou2023inf, herau20243dgs}. 
These studies solve the domain gap between LiDAR and camera by aligning the geometry information measured from both sensors.
However, the robustness of these methods depends on the differentiability of the underlying representation, and their accuracy is influenced by the rendering quality; thus, further improvement are required.

\begin{figure}[t]
    \centering
    \includegraphics[width=\linewidth]{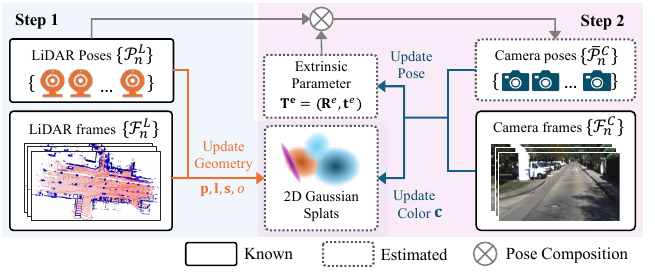}
    \vspace{-0.7cm}
    \caption{Overview of the proposed method. The proposed method uses LiDAR frames to reconstruct the geometric properties of 2D Gaussian splatting and optimizes the LiDAR-camera extrinsic parameters while updating the colors of the 2D Gaussian splats.}
    \label{fig:overview}
\end{figure}
Thus, we propose a LiDAR-camera calibration method to enhance robustness and accuracy using 2D Gaussian Splatting (2DGS), which leverages differentiability and high-quality rendering~\cite{huang20242dgs}. 
The proposed method is based on the concept of Implicit Neural Fusion (INF)~\cite{zhou2023inf}, which aligns 3D scenes constructed by LiDAR scans with the geometry information inferred from the camera images.
The proposed method begins by applying 2DGS to the LiDAR frames and then refines the Gaussian colors through photometric loss backpropagation from the camera images (Fig.~\ref{fig:overview}), thereby enabling the simultaneous optimization of the LiDAR-camera extrinsic parameters. 
Meanwhile, we identified certain limitations in relying exclusively on photometric loss; thus, we introduced reprojection and triangulation loss terms to overcome these issues.

Our primary contributions are summarized as follows:  \begin{itemize} 
    \item We propose a targetless LiDAR-camera calibration method that consolidates various geometric constraints based solely on the 2DGS representation. 
    \item We provide a mathematical analysis of the limitations inherent to the 2DGS for calibration
    and propose corresponding loss functions to address these limitations.
    \item We propose a depth weight uncertainty during the 2DGS reconstruction process to enhance the accuracy. 
\end{itemize}
We conduct experiments using the KITTI odometry dataset~\cite{Geiger2012CVPR}, to demonstrate the robustness and accuracy of the proposed method.

\section{RELATED WORK}
This section briefly reviews LiDAR-camera calibration methods, 3D and 2D Gaussian splatting techniques, 
and simultaneous pose estimation algorithms used in 3D reconstruction process.

\subsection{Targetless LiDAR-camera Calibration Methods}
Conventional appearance-based methods estimate relative poses using corresponding visual features.
The most practical and widely used methods ~\cite{levinson2013automatic, yuan2021pixel, kang2020automatic, zhang2021line} align image edges with geometric edges in LiDAR data. 
\cite{pandey2012automatic} and \cite{koide2023general} focused on maximizing the mutual information of intensity values between camera images and projected LiDAR points. 
In some studies~\cite{iyer2018calibnet, lv2021lccnet, shi2020calibrcnn, schneider2017regnet}, neural networks were employed for feature matching. 
The above methods have certain limitations:
the domain gaps between the 2D and 3D features reduce the accuracy and robustness, neural network (NN)-based methods require prior knowledge, and the capability to generalize NN-based methods remains unclear.

Motion-based methods~\cite{taylor2016motion, ishikawa2018lidar} rely on hand-eye calibration and estimate the extrinsic parameters by aligning the motion patterns of both sensors.
These approaches require each sensor to independently estimate the relative poses across multiple frames.
Additionally, motion-based methods depend on particular sensor movements to solve linear systems, which can be a significant limitation for certain robotic systems. 

To address these limitations, geometry-based methods have been increasingly investigated~\cite{chien2016visual, ishikawa2024lidar}. 
These approaches minimize the reprojection error to align LiDAR data with the geometry of multiview stereo cameras. 
Recent techniques have incorporated LiDAR-camera calibration within the process of 3D reconstruction~\cite{zhou2023inf, herau2023moisst} using NeRF~\cite{mildenhall2020nerf}. 
These methods bridge the domain gap by using the geometric consistency among sensors; however, their performance is highly dependent on the quality of the 3D reconstruction.
The proposed method aligns with this line of research and employs 2DGS for scene representation.

\subsection{Pose Estimation with Differentiable 3D Representations} 

3D reconstruction methods based on differentiable representations, such as NeRF, can estimate or improve camera poses in the optimization process.
If we have sufficient input images, we can determine the shape and camera poses simultaneously according to the Structure from Motion (SfM) theorem~\cite{ullman1979interpretation}. 
This theorem can also be applied to NeRF; thus, there are various methods to estimate camera poses~\cite{lin2021barf, bian2023nope, xie2024gfr}.
In addition, by leveraging consistent geometric information across sensors, some studies have achieved extrinsic calibration between LiDAR and cameras~\cite{zhou2023inf, herau2023moisst, li2024implicit}.
However, NeRF often faces challenges related to insufficient reconstruction quality, resulting in low calibration accuracy.

3DGS and the following 2DGS overcome these limitations by using explicit 3D Gaussian splats for scene representation.
While Gaussian splatting is not globally continuous like implicit neural representations (e.g., NeRF), it maintains local continuity and differentiability.
However, existing works have not fully leveraged these advantageous properties for pose estimation.
Instead, current approaches simply treat Gaussian splats as discrete points: Jiang et al.~\cite{jiang20243dgs} and Sun et al.~\cite{sun2024mm3dgs} apply traditional Perspective-n-Point (PnP) methods, while Herau et al. \cite{herau20243dgs}  use a hash-encoded MLP \cite{muller2022instant} to generate 3DGS properties.
While these methods have shown promising results, they primarily treat Gaussian splats as point features, leaving the potential benefits of their local continuity yet to be explored.

The proposed method leverages 2DGS because the precision and efficiency of Gaussian splatting can significantly improve the accuracy and speed of the calibration process. 
Additionally, we demonstrate that the continuity of Gaussian splatting can be effectively used by introducing geometric constraints, which is detailed in Sec.~\ref{sec:color_opt}.

\section{OVERVIEW and PRELIMINARY}
\label{sec:method1}
Our objective was to estimate LiDAR-camera extrinsic parameters throughout the 3D reconstruction process. 
In this section, we first outline the problem definition and provide an overview of the proposed method. 
Then, we introduce 2DGS as preliminary knowledge. 

\begin{figure*}[t]
    \centering
    \includegraphics[width=0.98\linewidth]{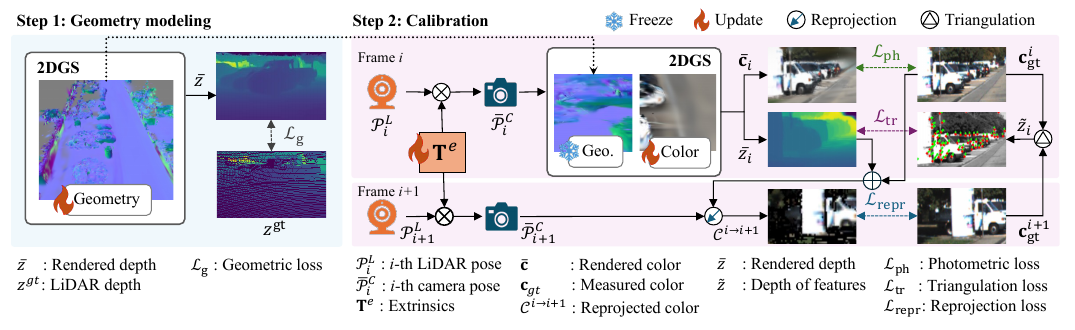}
    \vspace{-0.3cm}
    \caption{Workflow of the method. The right panel illustrates how the LiDAR frames were used to supervise the geometric properties of the 2DGS. As shown in the left panel, we freeze the geometric properties and update only the color properties during the calibration process. In addition to photometric loss, we also employ two interframe losses: triangulation loss and reprojection loss.}
    \label{fig:workflow}
    \vspace{-0.3cm}
\end{figure*}
%
%
\subsection{Problem Definition and Overview}
As illustrated in Fig.~\ref{fig:workflow}
, we use multiple sequential LiDAR frames $\{\mathcal{F}_1^\mathrm{L}, \mathcal{F}_2^\mathrm{L}, \dots, \mathcal{F}_N^\mathrm{L}\}$ with known corresponding poses $\{\mathcal{P}_1^\mathrm{L}, \mathcal{P}_2^\mathrm{L}, \dots, \mathcal{P}_N^\mathrm{L}\}$ and camera frames $\{\mathcal{F}_1^\mathrm{C}, \mathcal{F}_2^\mathrm{C}, \dots, \mathcal{F}_N^\mathrm{C}\}$ captured simultaneously as the input, where  $N$ is the number of frames. 
The LiDAR and camera are mounted on a rigid joint, which ensures that the extrinsic parameters between them remain consistent across all sequences. 
We represent the LiDAR-to-camera pose matrix as $\mathbf{T}^e\in \mathbb{R}^{4\times4}$, defined as follows:
\begin{equation}
\mathbf{T}^e = \begin{pmatrix}
\mathbf{R}^e & \mathbf{t}^e\\
\mathbf{0}_{1\times3} & 1
\end{pmatrix},
\end{equation}
where $\mathbf{R}^e \in \mathbb{R}^{3\times3}$ is the extrinsic rotation matrix and $\mathbf{t}^e \in \mathbb{R}^{3}$ is the extrinsic translation vector. 
An initial rough estimate of $\mathbf{T}^e$ is provided as input, with rotation parameters that allow partial overlap between the LiDAR point cloud projections and the camera's field of view.
It will then be refined by our program. 
Additionally, we assumed that the scene does not contain dynamic objects because all frames were used to reconstruct a static 2DGS.

In the proposed approach, we begin by using LiDAR frames and their corresponding LiDAR poses to train 2DGS, and we optimize only the geometric parameters.
After optimizing these geometric parameters, we fix them and then use the camera frames to update the colors of the splats. 
Throughout this process, we simultaneously estimate the extrinsic parameters between LiDAR and the camera. 
Fig.~\ref{fig:workflow} illustrates the workflow of the proposed method. 

\subsection{Preliminary: 2DGS}
2DGS~\cite{huang20242dgs} represents a scene as a collection of 2D Gaussian splats with the following learnable parameters: 
center position $\mathbf{p}_k \in \mathbb{R}^3(1 < k \leq N_k)$, opacity $o_k \in [0, 1]$, principal tangential unit vectors $\mathbf{l}_{k}^u \in \mathbb{R}^3$, $\mathbf{l}_{k}^v \in \mathbb{R}^3$, and scale factors $s_{k}^u \in \mathbb{R}$, $s_{k}^v \in \mathbb{R}$, where 
${N_k}$ is the total number of splats.
We define a 2D Gaussian in a world space with the 2D coordinates $\mathbf{u}=(u, v)$ on the tangential plane as follows: 
\begin{equation}
P(\mathbf{u}) = 
\mathbf{p}_k + s_k^u \mathbf{l}_k^u u + s_k^v \mathbf{l}_k^v v.
\label{eq:plane}
\end{equation}
Each splat also has an RGB color $\mathbf{c}_k \in \mathbb{R}^3$. 
The alpha value at the point $\mathbf{u}$ is given by the Gaussian function $\mathcal{G}_k(\mathbf{u})$ and the opacity of the splat as follows:
\begin{align}
\alpha_k(\mathbf{u}) &= 
o_k \mathcal{G}_k(\mathbf{u}).
\end{align}

To render a 2D splat at a camera pose $\mathcal{P}^\mathrm{C} \in \mathbb{R}^{4\times4}$ with an intrinsic projection matrix ${\mathbf{K}}_{3\times3}$ expanded to a $4 \times 4$ matrix as ${\mathcal{K}} = \begin{pmatrix} {\mathbf{K}}_{3\times3} & \mathbf{0}_{3\times1} \\ \mathbf{0}_{1\times3} & 1 \end{pmatrix}$, the 2D Gaussian $P$ is projected into the camera screen space as follows:
\begin{equation}
P^\mathrm{C} (\mathbf{u}) ={ {\mathcal{K}} }\mathcal{P}^\mathrm{C} P(\mathbf{u}) = (xz, yz, z, 1)^\mathsf{T}. 
\label{eq:intersection}
\end{equation}
Here, $\mathbf{x} = (x, y)$ represents the 2D image coordinate, and $z$ is the depth in the camera space. 
Please refer to the original paper for the inverse-calculation $\mathbf{u}$ from $\mathbf{x}$: $\mathbf{u}(\mathbf{x})$. 
Finally, the color $\mathbf{c}$
at pixel $\mathbf{x}$ is rendered by alpha blending as follows:
\begin{align}
    \mathbf{c}(\mathbf{x}) &= \sum_{k=1}^{N_k^\ast} \omega_k\mathbf{c}_k,\\
     \omega_k &= \alpha_k (\mathbf{u}(\mathbf{x})) \prod_{j=1}^{k-1}\left(1 - \alpha_j (\mathbf{u}(\mathbf{x})) \right).
     \label{eq:blending}
\end{align}
%
$N_k^\ast$ denotes the total number of 2D splats intersected with the ray of $\mathbf{x}$.
2DGS optimizes the above learnable parameters using the loss between the rendered color and the color from the input image. 

\section{GEOMETRIC 2DGS from LiDAR FRAMES}

{Unlike SfM with cameras, LiDAR provides absolute scale and precise depth measurements.} 
Thus, we first optimize the geometric properties of 2DGS $\left(\mathbf{p}_k, \mathbf{l}^u_k, \mathbf{l}^v_k, s^u_k, s^v_k, o_k, \epsilon_k \right)_{k=1}^{{N_k}}$ using only the LiDAR frames $\{\mathcal{F}^L\}$. 
$\epsilon$ is an additional parameter representing the depth uncertainty, which is used in the photometric loss (Sec. \ref{sec:color_opt}). 
We assume that the corresponding LiDAR poses 
$\{\mathcal{P}^L\}$ 
are known, as there are many off-the-shelf point cloud registration methods available~\cite{wang2019deep, yang2020teaser}. 
We use the downsampled LiDAR points as the seeds of the Gaussian splats (see details in Sec. \ref{sec:exp_setting}) and 
introduce a depth-specific splats adaptation method to enhance the quality of surface representation and reduce number of unnecessary. 

\vspace{-0.05cm}
\subsection{Depth-supervised 2DGS}
\vspace{-0.1cm}
We optimize the geometric parameters of the 2DGS by using the geometric loss $\mathcal{L}_\mathrm{g}$, 
which is composed of the depth loss $\mathcal{L}_{\text{d}}$, depth uncertainty loss $\mathcal{L}_{\text{unc}}$, depth distortion loss  $\mathcal{L}_{\text{dist}}$, and normal consistency loss $\mathcal{L}_{\text{norm}}$:
%
\begin{align}
    &\left(\mathbf{p}_k, \mathbf{l}_k, s_k, o_k, \epsilon_k\right)_{k=1}^{{N_k}} = \argmin_{\left(\mathbf{p}, \mathbf{l}, s, o, \epsilon\right)} (\mathcal{L}_\text{g}),\\
    &\mathcal{L}_\text{g} = \mathcal{L}_{\text{d}} + \mathcal{L}_{\text{unc}} + \lambda_\mathrm{d} \mathcal{L}_{\text{dist}} + \lambda_\mathrm{n} \mathcal{L}_{\text{norm}},
        \label{equ:lgeometry}
\end{align}
%
%
%
where $\lambda_\mathrm{d}$ and $\lambda_\mathrm{n}$ are weight paramters.
The definitions of $\mathcal{L}_{\text{dist}}$ and $\mathcal{L}_{\text{norm}}$ follow \cite{huang20242dgs}.

\subsubsection{Depth loss}The depth loss is the difference between the LiDAR depth and the rendered depth from 2DGS. 
The rays were sampled at all LiDAR points. 
The ray is cast from the LiDAR position, and its depth is represented as
$z^{\mathrm{L}}_i$. 
The depth loss is defined using the L1 loss between the LiDAR depth and the rendered depth $\Bar{z}_i$:
\begin{equation}
    \mathcal{L}_{\text{d}} = \frac{1}{N_r}\sum_{i}^{N_r}|\Bar{z}_i - z^{\mathrm{L}}_i|,
\end{equation}
where ${N_r}$ is the total number of rays. 
The rendered depth $\Bar{z}$ for a LiDAR ray is calculated as the weighted sum of all intersected Gaussian splat depths~\cite{huang20242dgs} as follows:
\begin{equation}
    \Bar{z} = \sum_{k=1}^{N_k^\ast} \omega_k z_k/\left(\sum_k \omega_k\right),
\end{equation}
where $z_k$ is the 
{ray-splat intersection depth}
calculated by Eq. (\ref{eq:intersection}), 
and $\omega_k$ is the weight of each splat 
calculated by Eq. (\ref{eq:blending}). 

\subsubsection{Depth uncertainty loss}
\begin{figure}[t]
    \centering
    \vspace{0.2cm}
    \includegraphics[width=0.8\linewidth]{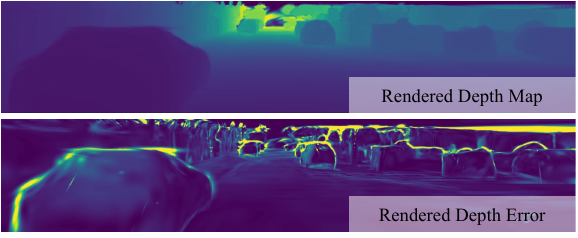}
    \caption{Example of the rendered depth error. A lighter color indicates a larger value. The depth error map emphasizes the edge components.}
    \label{fig:depth-error}
    \vspace{-0.5cm}
\end{figure}

%
To evaluate the errors caused by incorrectly defining the geometry of the 2DGS, we introduce depth uncertainty. 
The depth error for each ray can be rendered by alpha blending as follows:
\begin{equation}
    \bar{e_i} = 
    \sum_{k=1}^{N_k^\ast} \omega_k \epsilon_k .
\end{equation}
$(\epsilon_k)_{k=1}^{{N_k}}$ is updated based on the depth error of each ray, which is calculated as $e_i = |\bar{z}_i - z^{\mathrm{L}}_i|$. 
The uncertainty loss is described using the following L1 loss function:
\begin{equation}
    \mathcal{L}_{\text{unc}} = \frac{1}{{N_r}}\sum_i^{{N_r}} |\bar{e_i} - e_i|.
    \label{equ:lerror}
\end{equation}
Note that the gradient of $\mathcal{L}_{\text{unc}}$ will not be passed to $\alpha_k$.
Fig.~\ref{fig:depth-error} shows an example of the rendered depth error map. The depth errors are relatively large along the depth edges, which is consistent with our expectations.

\subsection{Splats Adaptation}
The original 3DGS and 2DGS splats were split and cloned based on the cumulative gradients (errors) of the existing splats.
This approach generates new splats only around existing splats and does not guarantee correct propagation of the isolated surfaces.

Thus, the proposed method directly places splats around LiDAR points according to the depth error.
If a LiDAR point is detected in a close range that differs significantly from the rendered depth ($\Bar{z}_i - z_i^{\mathrm{L}} > \theta_1(>0)$), we infer that there are insufficient splats near this LiDAR point; thus, we add a splat at that location. 
The initial surface normal of these splats aligns with the view direction.
Conversely, if the LiDAR point is located behind the rendered depth, the existing foreground splats will adjust accordingly; thus, no further action is required.


\section{LiDAR-CAMERA {CALIBRATION} with 2DGS}
\label{sec:color_opt}
After optimizing the geometric 2DGS, 
the colors $\mathbf{c}_k$ are estimated from the camera images. 
Because the camera poses are unknown, 
we jointly optimize the camera poses $\left(\Bar{\mathcal{P}}_i^C\right)_{i=1}^N$ with the colors. 
However, we only need to find the relative pose $\mathbf{T}^e$ between the camera and LiDAR because they are fixed. 
Thus, the poses of the camera frames $\left(\Bar{\mathcal{P}}_i^C\right)_{i=1}^N$ can be estimated with: 
    $\Bar{\mathcal{P}}^C_i = \mathbf{T}^e\cdot {\mathcal{P}}^L_i$
    \label{eq:pose}.
%

\subsection{Color and Pose Joint Optimization}
The joint optimization of $\mathbf{c}_k$ and $\mathbf{T}^e$ seeks to minimize the correspondence loss, $\mathcal{L}_\mathrm{c}$, which comprises the photometric loss $\mathcal{L}_{\text{ph}}$, triangulation loss $\mathcal{L}_{\text{tr}}$, and reprojection loss $\mathcal{L}_{\text{repr}}$:
%
\begin{align}
        \mathbf{T}^e, (\mathbf{c}_k)_{k=1}^{{N_k}} &=  \underset{\mathbf{T}^e, \{\mathbf{c}_k\}}{\mathrm{argmin}} \mathcal{L}_\mathrm{c},\\
    \mathcal{L}_\mathrm{c} &= \mathcal{L}_{\text{ph}} + \lambda_t \mathcal{L}_{\text{tr}} + \lambda_r \mathcal{L}_{\text{repr}}.
    \label{equ:lc}
\end{align}
Fig.~\ref{fig:workflow}
illustrates the optimization process. 
The photometric loss is defined as the L2 distance between the measured pixel color $\mathbf{c}^{\mathrm{L}}_i$ and the rendered color $\bar{\mathbf{c}}_i$ according to Eq. (\ref{eq:blending}). 
However, the geometric error influences the optimization of the relative pose $\mathbf{T}^e$; thus, we incorporate depth uncertainty weights as follows: 
\begin{align}
    \mathcal{L}_{\text{ph}} &=
        \frac{1}{\sum w_i}\sum_i^{{N_r}} w_i \left(\bar{\mathbf{c}}_i - \mathbf{c}^{\mathrm{L}}\right)^2,\\
    w_i &= \exp(-\bar{e}_i).
\end{align}

As described in~\cite{zhou2023inf} and~\cite{herau2023moisst}, the photometric loss optimizes $\mathbf{T}^e$ while updating $\{\mathbf{c}_k\}$.  
However, the photometric loss alone does not provide a sufficient update direction to effectively optimize the relative pose. 
Therefore, in the following sections, we closely examine the limitations of the photometric loss 
and propose solutions by incorporating the triangulation loss $\mathcal{L}_{\text{tr}}$ and the reprojection loss $\mathcal{L}_{\text{repr}}$. 

\subsection{Limitation of Photometric Loss}
\label{sec:limit}
Because we use the first-order gradient descent method for optimization, we can examine the update directions to $\mathbf{T}^e$ in each iteration by inspecting $\frac{\partial \mathcal{L}_{\text{ph}}}{\partial \mathbf{T}^e}$.
We denote the matrix dot product using the Frobenius inner product $\left<\cdot,\cdot\right>$~\cite{giles2008extended}. 
Refer to Appendix for the detailed derivation. 
%
By applying the chain rule of the derivatives, we can derive the following equation: 
\begin{equation}
    \frac{\partial \mathcal{L}_{\text{ph}}}{\partial \mathbf{T}^e_{[mn]}} \!=\! \sum_i^{N_k^\ast} \! \left< \! \frac{\partial \mathcal{L}_{\text{ph}}}{\partial u_i}, \frac{\partial u_i}{\partial \mathbf{T}^e_{[mn]}} \! \right> \!+\! \sum_i^{N_k^\ast} \! \left< \! \frac{\partial \mathcal{L}_{\text{ph}}}{\partial v_i}, \frac{\partial v_i}{\partial \mathbf{T}^e_{[mn]}} \! \right> \!,
    \label{eq:gradient}
\end{equation}
where ($u_i$, $v_i$) represents the intersected point with $i$-th
2D splat, as in Eq. (\ref{eq:intersection}). 
The subscript $(\cdot)_{[mn]}$ denotes the element at the $m$-th 
row and the $n$-th 
column.
We omit the factor $w_i/\sum w_i$ in the above equation because it is trivial for the analysis. 
\begin{table*}[t]
    \centering
    \vspace{0.3cm}
    \caption{Calibration results of the proposed method and comparision methods. Translation Error$^*$ is calculated with success cases.}
    \setlength\tabcolsep{4.5pt}
    \begin{tabular}{|c||c|c|c||c|c|c|}
    
    \hline
        & \multicolumn{3}{|c||}{{\bf Near}} &  \multicolumn{3}{|c|}{{\bf Far}} \\\cline{2-7}
       {\bf Method}& Success Rate $\%$& Rotation Error ($^\circ$)  & Translation Error$^*$ (cm) & Success Rate $\%$ & Rotation Error ($^\circ$)&Translation Error$^*$ (cm) \\\hline
        Edge~\cite{yuan2021pixel} & 0 & {127.96 $\pm$ 34.21} & --- & 0 & {119.14 $\pm$ 33.50} & ---\\\hline
        Intensity~\cite{koide2023general}&63& 1.07 $\pm$ 1.11&14.2 $\pm$ 9.1& 10 & {21.10 $\pm$ 32.42} & 61.2 $\pm$ 65.9 \\\hline
        Motion~\cite{ishikawa2018lidar} & 0 &129.79 $\pm$ 62.42  &--- & 0 & 131.18 $\pm$ 69.93 & --- \\\hline
        INF~\cite{zhou2023inf}\ &53 &5.74 $\pm$ 16.33 &15.3 $\pm$ 8.7 &23&6.17 $\pm$ 5.92 &20.0 $\pm$ 17.1\\\hline
        MOISST*~\cite{herau2023moisst}&10 &3.37 $\pm$ 2.37 &31.1 $\pm$ 14.0 & 0 & 11.78 $\pm$ 6.12 & ---\\\hline
        Ours w. Colmap & 97&0.39 $\pm$ 0.28 &9.9 $\pm$ 3.1 & 97 & 0.43 $\pm$ 0.45 & 10.2 $\pm$ 4.3 \\\hline
        Ours w. SS & \textbf{100} &  \textbf{0.36 $\pm$ 0.15} &\textbf{8.7 $\pm$ 3.2}  &\textbf{100}& \textbf{0.39 $\pm$ 0.18}& \textbf{8.8 $\pm$ 3.4 }\\\hline
        
    \end{tabular}
    \label{tab:calibration}
\end{table*}
\subsubsection{Robustness against initial pose error}
Because $\frac{\partial \mathcal{L}_{\text{ph}}}{\partial u_i}$ and $\frac{\partial \mathcal{L}_{\text{ph}}}{\partial v_i}$ are scalars, the spatial update directions are given by $\frac{\partial u_i}{\partial \mathbf{T}^e}$ and $\frac{\partial v_i}{\partial \mathbf{T}^e}$. 
Applying the chain rule, we derive the following equation: 
\begin{equation}
\begin{split}
            \frac{\partial \mathcal{L}_{\text{ph}}}{\partial \mathbf{T}^e_{[mn]}} &= \sum_i^{N_k^\ast}\left< \mathbf{h}_i
            , \frac{\partial \hat{\mathbf{p}}_{i}}{\partial \mathbf{T}^e_{[mn]}}\right>,\\
    \mathbf{h}_i &= \frac{\partial \mathcal{L}_{\text{ph}}}{\partial u_i}\cdot \frac{\mathbf{l}_{u_i}}{s_{u_i}} + \frac{\partial \mathcal{L}_{\text{ph}}}{\partial v_i}\cdot \frac{\mathbf{l}_{v_i}}{s_{v_i}},
            \label{eq:gradient-intersect}
\end{split}
\end{equation}
where $\hat{\mathbf{p}}_{i}$ represents the intersection of the 2D splat with the camera ray, relating to $u_i$ and $v_i$ as per Eq. (\ref{eq:plane}). 
This implies that each intersection point's update direction, denoted by $\mathbf{h}_i$, is restricted to its respective splat plane.
A key issue with this approach is that the gradients are limited to the local region within each splat’s size.
If the error in $\mathbf{T}^\mathrm{e}$ is large and the camera ray displacement significantly deviates, the update directions of the Gaussian splats may lose relevance.

We address the instability caused by local gradients by applying a reprojection loss (Sec.~\ref{sec:reprojection}), which directly retrieves the update direction from 2D image pixels rather than from 3D local spaces. 

\subsubsection{Instability in translation estimation}
Let us take a closer look at the extrinsic translation vector $\mathbf{t}^\mathrm{e}$. 
Here, we denote $\mathbf{n}_i$ as the normal vector of $i$-th 
Gaussian splat, $\mathbf{R}^\mathrm{C}$ as the rotation matrix of the estimated camera pose, and $\mathbf{r}$ as the ray direction in the camera space. 
We obtain the following equation: 
\begin{equation}
    \frac{\partial  \mathcal{L}_{\text{ph}}}{\partial \mathbf{t}^e_{[m]}} = \sum_i^{N_k^\ast} {\mathbf{h}_i}_{[m]} - \frac{\left<\mathbf{R}^C\mathbf{r}, \mathbf{h}_i \right>}{\left<\mathbf{R}^C\mathbf{r}, \mathbf{n}_i\right>}{\mathbf{n}_i}_{[m]},
    \label{eq:gradient-final}
\end{equation}
where $(\cdot)_{[m]}$ is the $m$-th 
element in the vector.

We find that the update direction of the camera pose includes components that are parallel and perpendicular to the splat surface.
However, the coefficient of the normal vector direction is affected by the camera ray direction. 
As described in \cite{zhang2014loam}, surfaces nearly parallel to camera rays are unreliable. 
Thus, valid Gaussian splats should ideally be angled relative to the view direction, minimizing the perpendicular component's coefficient in Eq. \ref{eq:gradient-final}. 
Nonetheless, as discussed in~\cite{ishikawa2024lidar}, the normal vector direction is critical for photometric loss, especially when there is minimal rotation in LiDAR-camera motion, such as in autonomous driving. 
In this case, the photometric loss alone is insufficient for accurate translation estimation.

Thus, we propose the use of the triangulation loss, which is described in detail in Sec.~\ref{sec:triangulation}.

\subsection{Reprojection Loss}
\label{sec:reprojection}
We incorporate the reprojection loss $\mathcal{L}_{\text{repr}}$ to enhance the robustness of the proposed method, particularly for optimizing the rotation component. 
Although errors in the ray direction can cause significant disparities in the intersected Gaussian splats, the reprojected pixels exhibit minimal displacement in the camera space, which facilitate accurate gradient computation. 
The displacement decreases further with increasing distance. 

For a pixel $\mathbf{v}_i \in \mathbb{R}^2$ in the $n$-th camera image, we reproject it to the $(n+1)$-th camera image to obtain the pixel coordinate $\mathbf{w}_i\in \mathbb{R}^2$ using the following equation:
\begin{equation}
    \begin{pmatrix}
            \mathbf{w}_i \\ 1
        \end{pmatrix}z_i^w= \bar{z_i}\mathbf{R}^{n\xrightarrow{}n+1}{\mathbf{K}}^{-1}\begin{pmatrix}
            \mathbf{v}_i \\ 1
        \end{pmatrix} + \mathbf{t}^{n\xrightarrow{}n+1},
        \label{eq:repr}
\end{equation}
where $\mathbf{R}^{n\xrightarrow{}n+1}$ and $ \mathbf{t}^{n\xrightarrow{}n+1}$ represent the rotation matrix and the translation vector from the $n$-th to the $(n+1)$-th camera space; $\bar{z}_i$ is the rendered depth value of pixel $\mathbf{v}_i$ from its original camera viewpoint, whereas $z_i^w$ is the reprojected depth. 
We then use a bilinear interpolation function $\mathcal{C}(\cdot)$ to approximate the color of $\mathbf{w}_i$ based on the four nearest pixels.
{To ensure robustness, we initially downsample the image pixels and progressively reduce the downsampling rate during training.} The reprojection loss is defined as follows:
\begin{equation}
        \mathcal{L}_\text{repr} = \frac{1}{{N_r}}\sum_i^{{N_r}} \left(\mathcal{C}\left(\mathbf{w}_i\right) - \mathcal{C}\left(\mathbf{v}_i\right)\right)^2.
\end{equation}

To handle occlusion when projecting from one camera view to another, we first render the depth maps from viewpoints $\mathbf{T}^C_1$ and $\mathbf{T}^C_2$. 
Then, we reproject the rendered depth from one view to another using Eq. (\ref{eq:repr}) and compare $z_i^w$ and the rendered depth $\bar{z}_i^w$ from the corresponding viewpoint. 
If the difference between these values exceeds a threshold value $\theta_2$, 
such points are excluded from the reprojection loss calculation.

\begin{figure*}[t]
    \centering
    \includegraphics[width=\linewidth]{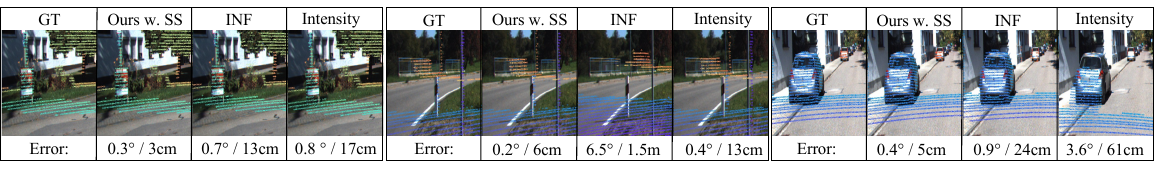}
    \vspace{-0.95cm}
    \caption{Ground truth image with LiDAR points projected using the resulting extrinsic parameters. An error of several degrees can cause significant displacement, whereas even an error of 20 cm does not result in noticeable displacement. Among the compared methods, the proposed method demonstrated robust results.}
    \label{fig:reproject}
\end{figure*}

\begin{figure*}[t]
    \centering
    \includegraphics[width=\linewidth]{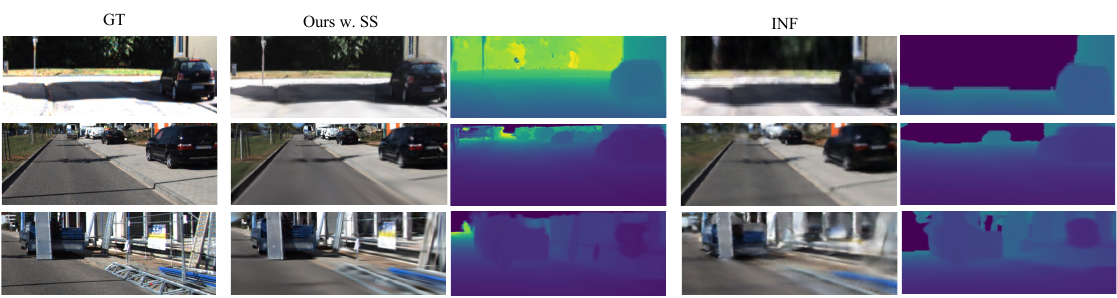}
    \caption{Rendered RGB and depth maps using proposed method and INF. The rendering quality of 2DGS (Ours) was much higher than that of MLP (INF).  Because volume rendering samples points within a certain range along a ray, points in distant areas are not sampled, resulting in those areas being unseen in the INF rendered image.}
    \label{fig:render}
\end{figure*}

\subsection{Triangulation Loss}
\label{sec:triangulation}
We propose applying the triangulation loss to the calibration with 2DGS. 
We assume that we can retrieve $M$ corresponding feature points in a sequential frames with coordinates $(\mathbf{q}_n^i)_{i=1}^M$ and $(\mathbf{q}_{n+1}^i)_{i=1}^M$, where each $\mathbf{q}_n^i, \mathbf{q}_{n+1}^i\in \mathbb{R}^2$ is from the $n$-th camera image and the $(n+1)$-th camera image, respectively. 
For simplicity, we omit the superscript $i$ in the following equations. 
Given $\mathbf{q}_n$ and $\mathbf{q}_{n+1}$, we can find the intersection point $\mathbf{Q}\in\mathbb{R}^3$ at which the rays from their respective camera positions in the world space intersect by
\begin{equation}
    \mathbf{Q} = \mathbf{T}_{n}^C{\mathbf{K}}^{-1}\begin{pmatrix}
        \mathbf{q}_{n}\\1
    \end{pmatrix}\Tilde{z}_n 
    = \mathbf{T}_{n+1}^C{\mathbf{K}}^{-1}\begin{pmatrix}
        \mathbf{q}_{n+1}\\1
    \end{pmatrix}\Tilde{z}_{n+1},
\end{equation}
where $\Tilde{z}_n$ and $\Tilde{z}_{n+1}$ are the two z-depths in the camera space from the camera position to $\mathbf{Q}$. 
 Denoting the $m$-th row of a matrix as $\cdot_{[m]}$, we can formulate as follows:
\begin{equation}
\frac{\mathbf{R}^{n\xrightarrow{}n+1}_{[1]}{\mathbf{K}}^{-1}\begin{pmatrix}
    \mathbf{q}_{n}
\\1\end{pmatrix}\Tilde{z}_n^{x} + \mathbf{t}^{n\xrightarrow{}n+1}_{[1]}}{\mathbf{R}^{n\xrightarrow{}n+1}_{[3]}{\mathbf{K}}^{-1}\begin{pmatrix}
    \mathbf{q}_{n}
\\1\end{pmatrix}\Tilde{z}_n^{x}+ \mathbf{t}^{n\xrightarrow{}n+1}_{[3]}} = ({\mathbf{K}}^{-1})_{[1]}\begin{pmatrix}
    \mathbf{q}_{n+1}
\\1\end{pmatrix}.
\end{equation}
We can derive $\Tilde{z}_n^{x}$ from the above equation, which provides an example of how to use the x-coordinates to determine the depth $\Tilde{z}_n$.
Similarly, we can formulate another equation by using the second row $\cdot_{[2]}$ instead of the first row $\cdot_{[1]}$ of the matrices to use the y-coordinates to calculate the z-depth, denoting as $\Tilde{z}_n^{y}$.

We describe the triangulation loss by comparing the calculated $\Tilde{z}_n$ with rendered depth $\Bar{z_i}$, as follows:
\begin{equation}
    \mathcal{L}_\text{tr} = \frac{1}{2}\cdot \frac{1}{{N_r}}\sum_i^{{N_r}} \left(\mathcal{T}\left(\Tilde{z}_n^{x}, \Bar{z_i}\right) + \mathcal{T}\left(\Tilde{z}_n^{y}, \Bar{z_i}\right)\right),
\end{equation}
where $\mathcal{T}(\cdot, \cdot)$ denotes the Tukey robust function, which mitigates the influence of outliers. 
The threshold of Tukey loss is set to 1~m in this experiment. 
Because $\Tilde{z}_n$ and $\Bar{z_i}$ are functions of the extrinsic parameters $\mathbf{T}^e$, the gradients are backpropagated through both quantities.

\section{EXPERIMENTS}

\subsection{Dataset and experimental settings}
\label{sec:exp_setting}
We used 30 sequences in the KITTI odometry dataset~\cite{geiger2012automatic}, each comprising 50 sequential LiDAR-camera pairs without dynamic objects. 
The extrinsic parameters in $\mathbf{T}^e$ are expressed using the Lie algebra in $\mathfrak{se}{3}$. 
To simulate different levels of the initial calibration error, we introduced a "far" initial bias by adding 0.2 to all the $\mathfrak{se}{3}$ parameters, resulting in an error of ${16.84}^\circ / 29.25cm$. 
{The scalar rotation error is defined as the magnitude of the error represented by the rotation vector.}
Additionally, we applied a "near" initial bias by adding 0.1 to the translation parameters while leaving the rotation parameters unchanged, leading to an error of $0^\circ /14.68cm$.
{On an RTX4090 GPU, our method required 2.0 and 6.8 minutes for geometry optimization and calibration, respectively.}
{For more results on other dataset, please refer to the Appendix.}

In the process of generating geometric Gaussian splats from the LiDAR frames, we begin by separating the ground points from nonground points. 
The ground points were downsampled using a 0.5~m voxel grid, while the remaining points were downsampled using a finer 0.15~m voxel grid. 
These downsampled points serve as the initial Gaussian splats. 
We densify the points by cloning them four times in the final RGB rendering. 
We set the threshold value $\theta_1$ for splats adaptation to 0.5~m and $\theta_2$ for occlusion handling to 0.05~m. 
We set weight parameters $\lambda_\mathrm{d}$, $\lambda_\mathrm{n}$, $\lambda_t$, and $\lambda_r$ to $1e^{4}$, $1e^{-1}$, $1$, and $200$, respectively.

\begin{figure*}[t]
    \centering
    \includegraphics[width=\linewidth]{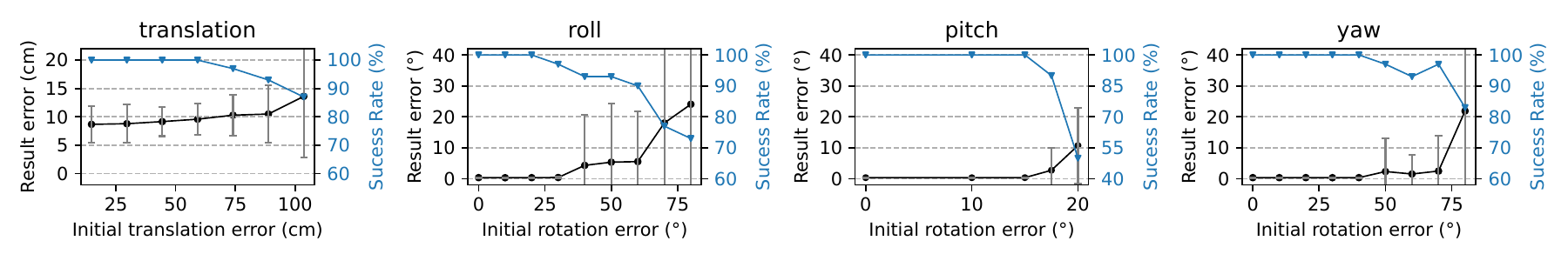}
    \includegraphics[width=0.9\linewidth]{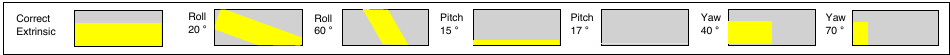}
    \caption{Success rate and calibration error with respect to different initial biases. Here, we illustrate how the target LiDAR point cloud is reprojected onto the camera image relative to the initial rotation error: the gray background denotes the camera image area, and the yellow foreground denotes the target LiDAR area. The proposed method does not work if the target LiDAR points are initially not visible. }
    \label{fig:convergence}
\end{figure*}
We employed the Adam optimizer~\cite{kingma2014adam} for the optimization process. 
The learning rates of the Gaussian properties were set in accordance with the original 2DGS ~\cite{huang20242dgs}. 
To address the sensitivity of the translation parameters to the initial rotation errors, we initialized the learning rate for the rotation parameters at $1e^{-2}$ and for the translation parameters at $5e^{-4}$. 
Once the rotation change was less than $0.1^\circ$ over the last 500 iterations, we adjusted the learning rates, setting the rotation to $1e^{-3}$ and the translation to $1e^{-2}$. 
The learning rate is halved when the change rate of translation falls below $1e^{-5}$ and $5e^{-6}$. 
We uses 15000 iterations for both the geometric 2DGS generation and joint optimization processes.

\subsection{Evaluation of Calibration Accuracy}
We present the results of our calibration method using two feature extraction and matching approaches for triangulation loss $\mathcal{L}_{\text{tr}}$: Colmap features and matching, denoted as "Ours w. Colmap," and SuperPoint~\cite{detone2018superpoint} with SuperGlue~\cite{sarlin2020superglue}, denoted as "Ours w. SS." 
To evaluate calibration accuracy and robustness, we compared our method with the following approaches:
\begin{itemize}
    \item Edge~\cite{yuan2021pixel}: Align the projected LiDAR edges with the camera edges.
    \item Intensity~\cite{koide2023general}: Maximize the mutual information between the reflectivity of the LiDAR scans and the intensity of the camera images. 
    In the "near" experiment, we only applied the refinement process described in~\cite{koide2023general}; in the "far" experiment, we applied the automatic initial guess in~\cite{koide2023general}. 
    \item Motion~\cite{ishikawa2018lidar}: Estimate camera motion with LiDAR point cloud and extrinsic parameters using hand-eye calibration.
    \item INF~\cite{zhou2023inf}: Optimize the pose during NeRF~\cite{mildenhall2020nerf} reconstruction.
    \item MOISST$\divideontimes$~\cite{herau2023moisst}: Optimize the pose during InstantNGP~\cite{muller2022instant} reconstruction. Implemented ourselves.
\end{itemize}
Because rotation errors can heavily affect translation, we considered results with rotation errors under $1^\circ$ as successful and excluded failures from the translation error calculation. 
$^\ast$ indicates errors calculated only for successful cases.  
Thus, the translation and rotation errors represent accuracy, whereas the success rate reflects robustness. 
Table \ref{tab:calibration} provides the detailed results, and Fig.~\ref{fig:reproject}. shows a qualitative reprojected example. 

The results demonstrate that the proposed method outperforms existing approaches. 
Appearance-based methods face challenges with the domain gap between LiDAR and camera RGB data because LiDAR reflectivity does not always correlate with RGB intensity, and depth or normal edges in LiDAR scans differ from texture edges in RGB images.  
Additionally, the KITTI dataset is noisy and lacks structure; thus, edge-based methods are difficult to extract meaningful features.
Motion-based calibration requires rotational motion in multiple directions, which hinders convergence when the data includes large linear motions. 
INF~\cite{zhou2023inf} exhibits low reconstruction quality (see Fig.~\ref{fig:render}), while hash encoding methods like ~\cite{herau2023moisst} struggle with large initial rotation errors. 
In contrast, the proposed method demonstrate robust performance even with initial errors, and the translation errors remained within acceptable limits.

\subsection{Ablation Study on Losses}
In this section, we demonstrate the effectiveness of the triangulation loss, reprojection loss, and depth uncertainty weights. 
We conducted ablation studies by removing each of these components in the optimization process, starting with the "far" initial extrinsic parameters. 
Table \ref{tab:ablation} lists the results. 
The rows, from top to bottom, represent the calibration results for: our method without triangulation loss, using Colmap feature points for triangulation loss, without reprojection loss, without depth uncertainty weights, and using SuperPoint-SuperGlue feature points for triangulation loss.

We can notice that the results highlight the critical role of reprojection loss, where its removal leads to a significant drop in robustness, reducing the success rate to $30\%$. 
The triangulation loss and depth uncertainty weights improve the accuracy by minimizing the translation errors. 
Interestingly, the translation error remained largely unchanged whether the Colmap-matched points are used or the triangulation loss is entirely omitted. 
This outcome aligns with the analysis in Sec.~\ref{sec:limit}, where the focus is on points on surfaces, whereas the Colmap features are predominantly edge points, which are often filtered out by the Tukey loss due to the instability of the depth in edges. 
{Additionally, we conducted experiments using 3DGS as the base representation, with detailed results presented in the Appendix.}

\begin{table}[t]
    \centering
    \caption{Ablation experiment results with different experiment settings.}
    \vspace{-0.2cm}
    \begin{tabular}{|c|c|c|c|}
    \hline
       \multirow{ 2}{*}{Method}& Success &Rotation & Translation\\
       & Rate $\%$  & Error ($^\circ$)& Error$^*$ (cm) \\\hline
       Ours wo. triangulation & 97 & 0.44$\pm$0.38 & 11.18$\pm$3.4\\\hline
       Ours colmap &97& 0.44 $\pm$ 0.45 & 10.16 $\pm$ 4.34 \\
       \hline
       Ours wo. reprojection & 30  & {9.28$\pm$12.06}& 11.67$\pm$3.14\\\hline
       Ours wo. weights & \textbf{100}& 0.40$\pm$0.16 & 9.93$\pm$4.22 \\\hline

       Ours full &\textbf{100}& \textbf{0.39 $\pm$ 0.18}& \textbf{8.79 $\pm$ 3.40} \\
       \hline
        
    \end{tabular}
    \label{tab:ablation}
\end{table}

\subsection{Evaluation of the Calibration Robustness}
To account for the sensitivity of the initial rotation error to the axes, we separately evaluated the impact of the initial rotation errors along each axis on the calibrated rotation parameters. 
For rotation errors, we define success cases as those with an error within $1^\circ$, as in the previous experiments.
For the translation results, errors within 20 cm were considered to be successful. 
Fig.~\ref{fig:convergence} shows the results. 
In our experiment, our method successfully converged to the correct pose from the initial errors of $20^\circ$ (roll), $15^\circ$ (pitch), $40^\circ$ (yaw), and 60~cm in translation. 
In addition, we achieved a 90$\%$ success rate for initial errors within $60^\circ$ (roll), $17^\circ$ (pitch), $70^\circ$ (yaw), and 80~cm in translation.
\section{CONCLUSIONS}
This paper presents a LiDAR-camera calibration method that uses 2D Gaussian Splatting (2DGS). 
The proposed method begins by constructing the 2DGS geometry from the LiDAR scans, thereby creating a geometric scene representation. 
The calibration is then integrated into the 2DGS colorization stage to align the LiDAR data with the captured camera images. 
We analyzed the limitations of 2DGS in the pose estimation task and addressed them by introducing triangulation and reprojection losses, along with a depth uncertainty weighting scheme to enhance the calibration stability. 
We validated the proposed method on the challenging KITTI odometry dataset.  
The results demonstrated that the proposed method improves alignment accuracy and robustness, outperforming existing methods in terms of precision and handling of complex scenarios.

\bibliographystyle{IEEEtran}
\bibliography{myref}
\clearpage

\end{document}